\documentclass[journal]{IEEEtran}
\usepackage{filecontents}
\usepackage{graphicx}
\usepackage{multicol}
\usepackage{subfig}
\usepackage{amsmath}
\usepackage{amsfonts}
\usepackage{romannum}
\usepackage{caption}
\usepackage{tabularx}
\usepackage{tabulary}
\usepackage{array}
\usepackage{tabu}
\usepackage{mathtools}
\usepackage{amssymb}
\usepackage{tikz}
\usepackage{pgfplots}

\ifCLASSINFOpdf
\else
\fi

\begin{document}

\title{Voxel-level Siamese Representation Learning for Abdominal Multi-Organ Segmentation}
\author{Chae Eun Lee, Minyoung Chung$^{\ast}$, and Yeong-Gil Shin%
\thanks{\textit{Asterisk indicates corresponding author.}}%
\thanks{C. Lee, and Y.-G. Shin are with the Department of Computer Science and Engineering, Seoul National University, South Korea (e-mail: nuguziii@cglab.snu.ac.kr).}%
\thanks{*M. Chung is with the School of Software, Soongsil University, South Korea (e-mail: chungmy@ssu.ac.kr).}
}


\maketitle

\begin{abstract}
Recent works in medical image segmentation have actively explored various deep learning architectures or objective functions to encode high-level features from volumetric data owing to limited image annotations. However, most existing approaches tend to ignore cross-volume global context and define context relations in the decision space. In this work, we propose a novel voxel-level Siamese representation learning method for abdominal multi-organ segmentation to improve representation space. The proposed method enforces voxel-wise feature relations in the representation space for leveraging limited datasets more comprehensively to achieve better performance. Inspired by recent progress in contrastive learning, we suppressed voxel-wise relations from the same class to be projected to the same point without using negative samples. Moreover, we introduce a multi-resolution context aggregation method that aggregates features from multiple hidden layers, which encodes both the global and local contexts for segmentation. Our experiments on the multi-organ dataset outperformed the existing approaches by 2\% in Dice score coefficient. The qualitative visualizations of the representation spaces demonstrate that the improvements were gained primarily by a disentangled feature space.
\end{abstract}


\IEEEpeerreviewmaketitle

\section{Introduction}

\IEEEPARstart{M}{ulti}-organ segmentation of organs such as the liver, pancreas and stomach in abdominal computed tomography (CT) scans is an essential task for computer-aided diagnosis (CAD), image-guided robotic surgeries, radiation therapy, and virtual surgeries \cite{cad, robotics_surgery, 788580, ma2020abdomenct1k}. Accurate and reliable segmentation results are required to optimize clinical workflow, such as the planning of surgeries or treatments. With the universal adaptability of convolutional neural networks (CNNs) \cite{10.5555/2999134.2999257}, different levels of visual features and patterns of organs in medical images can be efficiently detected and trained. Yet, a large amount of annotated data is required to accelerate a CNN to accurately detect medical imaging findings and precisely segment the organs \cite{cho2016data}, which is considered as an expensive and time-consuming task. Thus, the embedding of rich semantic information by exploiting limited data is a primary challenge in medical image segmentation task.

\begin{figure}[t]
    \centering
    \includegraphics[width=1\linewidth]{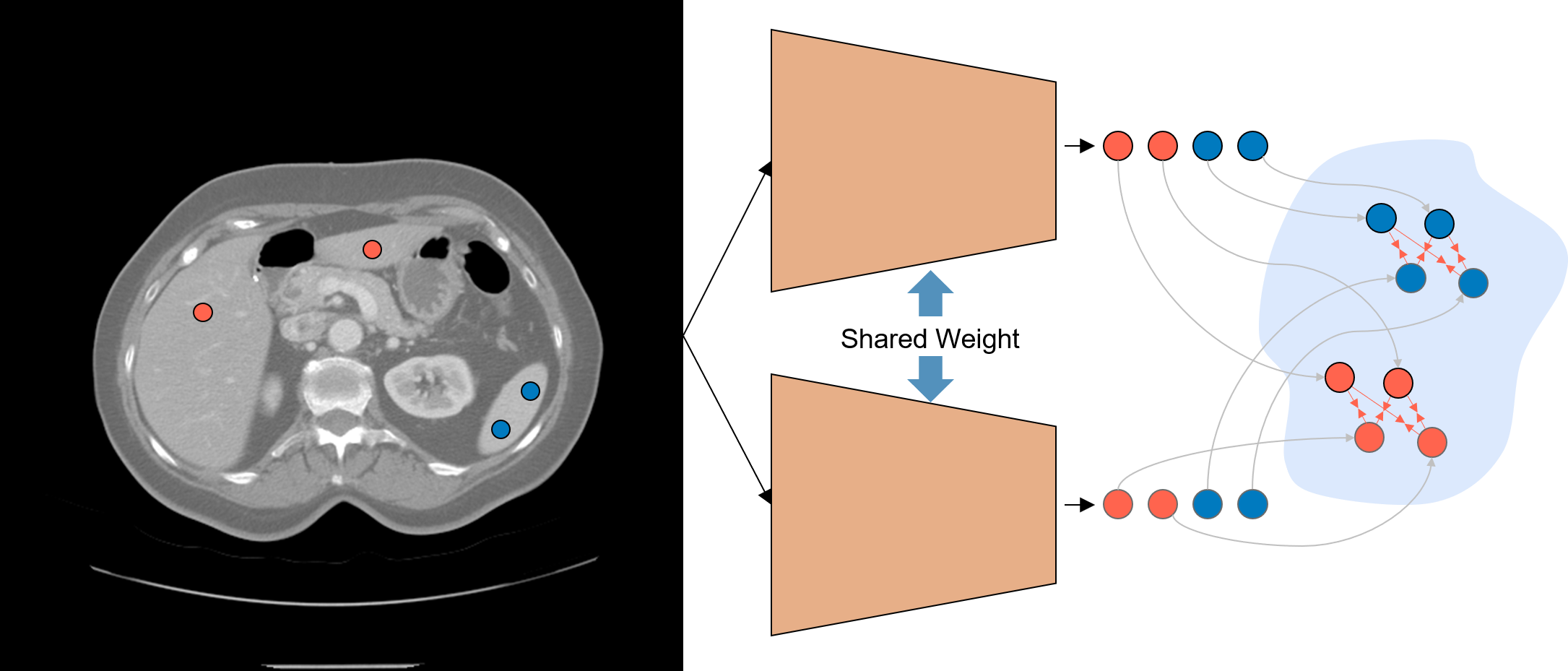}
    \caption{\textbf{Our proposed method} Orange and blue dots on the computed tomography scan refers to voxels; the voxels with the same color are from the same class. Two sets of voxel-level features are processed by the same network. Then, each of the features in the same class and from different networks are pulled closer to each other, thereby maximizing similarity (sky blue-colored area is an embedding space).}
    \label{fig:summary}
\end{figure}

Recently, efforts have been devoted to improve the performance of segmentation based on encoding high-level features, which is also known as feature learning \cite{7954012}. Different types of architectures have been developed to increase the representation power of volumetric CT images \cite{VoxelResNet, 3DUNet, ResidualUNet, VNet}. Various objective functions, such as distributed-based losses \cite{celoss, focal_loss}, region-based losses \cite{dice_loss, log_dice}, and boundary-based losses \cite{hausdorff} have been proposed for domain-specific problems. Despite their improved representations for 3D medical data, these methods focus on the local context instead of the global context for dense segmentation results, which indicates that these methods are prone to be affected by data distribution. Attention-gated mechanisms \cite{attention_1} were proposed for fusing both global and local contexts, learning only relevant regions in input images, or leveraging contextual information over multiple slices \cite{attention_2}. However, these methods ignore the cross-image global context and define the context relations in the final feature space (i.e., the last feature layer or decision space), which significantly reduces the representation power of the intermediate hidden layers. Meanwhile, self-supervised learning methods based on contrastive loss \cite{8578491, oord2019representation, simclr, moco}, which defines the context relations in the representation space, have proven the effectiveness of learning representations. Recently, contrastive loss was employed in semantic segmentation tasks in the natural images \cite{wang2021exploring}, which demonstrated the effectiveness of contrastive loss for learning the global context in a supervised learning framework. However, these methods used memory banks or large batch sizes for a large number of negative samples, which resulted in a high computational cost. Furthermore, embedding the local context along with the global context is essential for accurate segmentation, which was not addressed in \cite{wang2021exploring}.

Inspired by recent methods of representation learning in a self-supervised learning framework, we aim to improve the representation power for multi-organ segmentation task. We can define the problem of multi-organ segmentation on CT scans as a voxel-wise classification problem; consequently, the voxel-wise semantic information must be embedded for both global and local context. However, there are two limitations that occur when applying previous methods to volumetric CT images segmentation. \textbf{(1)} Owing to the extensive requirement of large negative samples, when employing contrastive loss, the current methods result in additional computational time and cost, which are primarily increased by the size of the volume data. \textbf{(2)} The current methods only have an advantage in the global context, not the local context. This is because these methods only take advantage of the last layer for feature embedding.

To resolve the aforementioned issues, we propose effective voxel-level representation learning method for multi-organ segmentation. Based on a previous research \cite{wang2021exploring}, we defined positive samples as voxels from the same class (organs) and computed voxel-to-voxel and voxel-to-region contrasts that enforce embeddings to be similar for positive samples. The uniqueness of our method is that we \textbf{do not employ negative samples}, which is the solution for the first limitation \textbf{(1)}. Thus, computational cost can be significantly reduced because we can work with a typical batch size \cite{simclr} and do not require additional memory bank \cite{byol} for negative (voxel) samples. Another recent work, i.e., \textbf{SimSiam} \cite{simsiam}, proposed a simple Siamese network and stop-gradient operation, and further proved that neither negative samples nor momentum encoders are critical for the comparative performance. Thus, we adopted the SimSiam method as a baseline for our representation learning and customized it for our supervised voxel-level representation embedding. We build our representation learning network using a standard 3D U-Net architecture. The 3D U-Nets \cite{3DUNet} are commonly used for image segmentation tasks because of their good performance and efficient use of GPU memory \cite{attention_1}. For the solution for the second limitation \textbf{(2)}, we propose a \textbf{multi-resolution context aggregation} in the 3D U-Net architecture for embedding both local and global contexts, which enables both semantic information and precise localization.

To summarize, our contributions are as follows:

\begin{itemize}
  \item We propose simple yet effective \textbf{voxel-level representation learning} method for multi-organ segmentation on abdominal CT scans. Our method enforces voxel-level feature relations in the representation space so that we can enhance representation power of the base network (i.e., 3D U-Net \cite{3DUNet}).
  \item We define voxel-level feature relations \textbf{without using negative samples}, which is an efficient method in terms of the computational cost. While using SimSiam method \cite{simsiam}, we neither use a large batch size nor use a momentum encoder, which are typically required for collecting a large amount of negative samples.
  \item We propose a \textbf{multi-resolution context aggregation} method that aggregates features from the intermediate layers and the last hidden layer. Using our method, we can train both global and local context features simultaneously.
  \item Our proposed method shows superior performance when compared to the existing state-of-the-art methods. Moreover, our method can be effortlessly combined with any base network without extra parameters during inference. Furthermore, we demonstrate that our method is effective with a limited dataset.
\end{itemize}

\par

\section{RELATED WORK}
\subsection{Medical Image Segmentation} 

The state-of-the-art models for medical segmentation are based on encoder-decoder architectures, such as U-Net \cite{UNet} (includes contracting path, expanding path, and skip connections). For dense segmentation results in volumetric images, encoder-decoder based architectures \cite{VoxelResNet, 3DUNet, ResidualUNet, VNet} using 3D convolutional networks were proposed, and the weighted cross-entropy loss \cite{celoss} or Dice coefficient \cite{dice_loss} was used as a basic loss function. In this setting, voxel-level feature learning can embed a rich local context that enables precise localization. However, it is limited to the local context, and it is difficult to capture long-range dependencies. To overcome this limitation, attention-gated mechanisms \cite{attention_1, attention_2} were proposed, which are considered to be efficient for fusing both global and local semantic information. Howevere, previous attention mechanisms have a common limitation that it is difficult to represent context relations in the representation space.

\subsection{Contrastive Learning}

Self-supervised learning methods based on contrastive loss \cite{1640964, 8578491, oord2019representation, hjelm2019learning, h2020dataefficient, simclr, moco} have proven the effectiveness of learning representations that are invariant to different views of the same instance. In contrastive learning, images from multiple (similar) views are pulled closer together, while all other images (negative samples) are pushed away. Defining the positive (similar) and negative samples is an essential part of contrastive learning, and in general, data augmentation of the same image is used for generating positive samples, and different images are used as negative samples. It is commonly known that contrastive loss benefits from more negative examples \cite{1640964, tian2020contrastive} (collapsing occurs without negative samples). Thus, previous studies relied on large batch sizes \cite{simclr} or memory banks \cite{moco}. Beyond contrastive learning, recent studies \cite{byol, simsiam} have shown promising results for representation learning by employing only positive pairs without using the negative samples. A momentum encoder and moving-average updating technique \cite{byol} or the SimSiam method \cite{simsiam} are well-known techniques that do not use negative pairs to prevent the occurrence of a collapsing. In this work, we used SimSiam \cite{simsiam} as a baseline of our feature learning network to capitalize on the computational efficiency by using only positive pairs.

\begin{figure*}[h!bt]
    \centering
    \includegraphics[width=\linewidth]{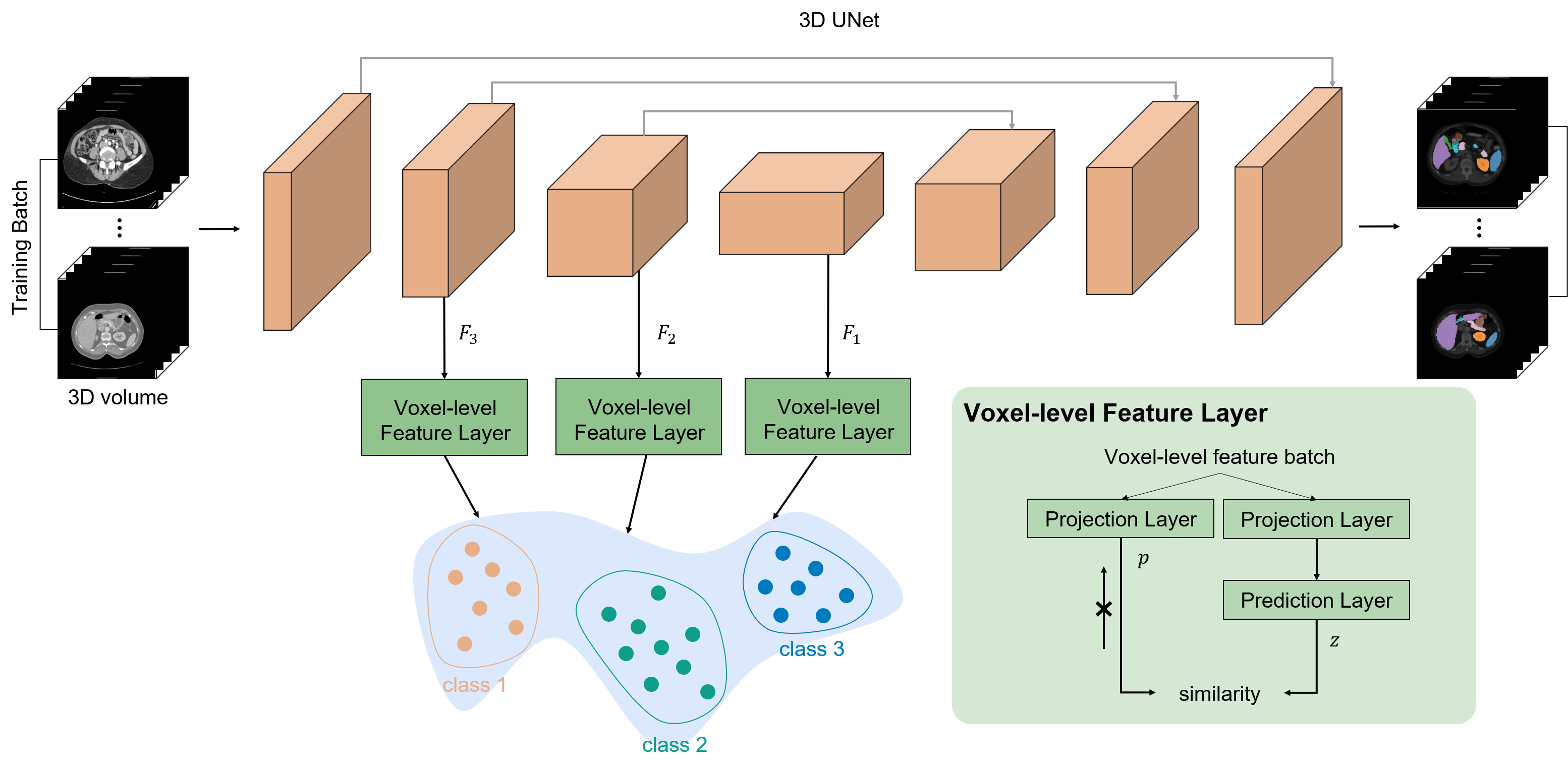}
    \caption{\textbf{Overview of the proposed architecture} Voxel-level feature layer takes features of hidden layers as input and defines voxel-level feature relations in the representation space (sky blue-colored area). Using a Siamese network (encoder and projection layer), voxel-level feature batch ($F_i$) is projected to $p$. Similarity between $p$ and $z$ is minimized by stop-gradient technique.}
    \label{fig:architecture}
\end{figure*}

Furthermore, contrastive learning has been applied to medical or natural image segmentation to enhance the representation power. There are certain recent works \cite{chaitanya2020contrastive, taleb20203d} that used contrastive loss for the pretraining encoder, which was beneficial for the downstream medical segmentation task. Moreover, another current work \cite{wang2021exploring} took advantage of contrastive loss in a supervised learning framework and showed outperforming results in natural image semantic segmentation tasks. This work embedded global context using pixel-to-pixel and pixel-to-region contrasts, thereby utilizing the annotated labels of each pixel. Inspired by this work \cite{wang2021exploring}, we propose a method for voxel-level representation learning that embeds both local and global contexts.

\section{PROPOSED METHOD}

\subsection{Preliminaries}

The core idea of visual self-supervised learning is to train the encoder to embed high-level visual features from the data itself. To achieve this, simple Siamese (SimSiam \cite{simsiam}) method takes two randomly augmented (i.e., rotation, color jitter or scaling) views $x_1$ and $x_2$ as an inputs to the encoder and minimizes negative cosine similarity of the two output vectors, $p_1$ and $z_2$:
\begin{equation}
    \mathcal{D}(p_1, z_2) = - {p_1 \over \lVert p_1 \rVert_2} \cdot {z_2 \over \lVert z_2 \rVert_2},
\label{eq:1}
\end{equation}
where $\lVert \cdot \rVert_2$ is $l_2$-norm, $p_1 \triangleq h(k(f(x_1)))$ and $z_2 \triangleq k(f(x_2))$ (encoder network $f$, projection MLP head $k$, and prediction MLP head $h$.)

A key component in SimSiam is the stop-gradient, which prevents a network $f$ from collapsing. The total form of the SimSiam loss is as follows:
\begin{equation}
    \mathcal{L} = {1 \over 2}\mathcal{D}(p_1, stopgrad(z_2)) + {1 \over 2}\mathcal{D}(p_2, stopgrad(z_1)),
\label{eq:2}
\end{equation}
where $stopgrad$ indicates that the loss is not updated by the gradients of its parameters, which is PyTorch-like style.

\label{section:pre}

\subsection{Voxel-level Representation Learning}
\label{section:3-b}
Our architecture (Fig. \ref{fig:architecture}) is composed of a base network (3D U-Net) and voxel-level feature layers. We first demonstrate a basic 3D U-Net network training scheme and introduce our novel voxel-level representation learning method.

3D U-Net takes a 3D volume, $x \in \mathbb{R}^{H \times W \times D}$ as an input, and the encoder produces each features $F_i \in \mathbb{R}^{H \times W \times D \times C_i}$ from the $i^{th}$ layer (from the last). For each upsampling step in the decoder (expanding path), these features ($F_i$, except for the $1^{th}$ layer) are concatenated with the upsampled feature map, which is connected with the gray line in Fig. \ref{fig:architecture}, and is also known as skip connection, which is crucial for precise localization. Then, the last layer outputs the score map $\mathcal{S} \in \mathbb{R}^{H \times W \times D \times Class}$ for each class. The Dice coefficient is used in most 3D medical image segmentation tasks as a loss function; therefore, we used multi-label soft Dice loss \cite{VNet} to maximize the overlap between the ground truth and prediction. The multi-label soft Dice loss is defined as:
\begin{equation}
    \mathrm{L}_{dice} = \sum^{class}_c{{2\sum^{H \times W \times D}_i {p_i^c g_i^c}} \over {\sum_i^{H \times W \times D} {(p_i^c)}^2 + \sum_i^{H \times W \times D} {(g_i^c)}^2}},
    \label{eq:3}
\end{equation}
where $p_i \in \mathbb{R}^{H \times W \times D}$ is ground truth and $g_i \in$ \textit {softmax}$(\mathcal{S})$

A lot of work in medical imaging tasks uses Dice coefficient instead of cross-entropy as it works better for class imbalance. However, similar to the problem of cross-entropy loss as stated in \cite{wang2021exploring}, Dice loss also penalizes voxel-wise predictions independently, ignoring voxel-to-voxel relations. It is true that (\ref{eq:3}) is only computed for each predicted voxel to determine if it is in the ground-truth mask region. In the updating phase, the loss of each voxel prediction is backpropagated independently so that the network cannot learn their relations. Further, prediction is performed in the last layer of the decoder, and it is not sufficient to encode high-level features in the encoder. As shown in recent downstream tasks in medical imaging \cite{chaitanya2020contrastive, taleb20203d}, a pretraining encoder that uses pretext tasks or contrastive loss before training the entire segmentation network leads to substantial performance improvements, which demonstrates that there is room for performance improvement in the encoder.

To tackle these issues, additional loss for accelerating cross-voxel relations and encoding sufficient high-level features in the encoder is required. A recent paper \cite{wang2021exploring} proposed a pixel-wise contrastive algorithm and showed the result of a well-structured semantic feature space. In this study, we extended this method to a voxel-wise contrastive algorithm that defines the voxel-to-voxel and voxel-to-region contrasts. Thus, we can suppress voxel-level relations to obtain similar features from the same class. Furthermore, we defined voxel-wise relations without using negative samples, which is the main difference from the previous method \cite{wang2021exploring}. We used SimSiam method that utilizes both the Simaese network and the stop-gradient technique. We can reduce the computational cost while maintaining the competitive performance because we did not use either a large batch size or a memory bank for negative samples.

We propose a voxel-level feature layer that consists of two MLP heads, i.e., projection and prediction. The voxel-level feature layer takes $F_i$ as an input. The $F_i$ feature is converted into $C_i$-d vector with a batch size of $H \times W \times D$, which can be referred to as $(H \times W \times D) \times C_i$. Then, two identical $C_i$-d features, $f$ are passed through the projection layer and one of the outputs is passed through the prediction layer, which can be represented as $p \triangleq pred(proj(f))$ and $z \triangleq proj(f)$, where $pred$ is prediction layer and $proj$ is projection layer. Let the sets of every $p$ and $z$ from different voxels be represented as $\mathcal{P}$ and $\mathcal{Z}$, respectively, and let the sets of voxels from the same class $c$ be represented as $\mathcal{P}^c$ and $\mathcal{Z}^c$, respectively. Then, following (\ref{eq:1}), the voxel-wise loss function that maximizes the similarity from the same class can be defined as:
\begin{equation}
    \mathcal{L}_{voxel-wise}(p, z) = \sum_{p \in \mathcal{P}^c} \sum_{z \in \mathcal{Z}^c} {p \over \lVert p \rVert_2} \cdot  {z \over \lVert z \rVert_2}.
\label{eq:4}
\end{equation}
Then, to minimize the loss using stop-gradient technique for multi-class relations, the loss for training can be defined as:
\begin{equation}
   \mathcal{L} = - \sum_c^{class}\mathcal{L}_{voxel-wise}(p, stopgrad(z)).
   \label{eq:5}
\end{equation}
The loss function according to \ref{eq:5} suppresses voxels from the same class so that they can be projected onto the same point in the embedding space. Further, we can easily extend this loss to the intra-voxel-wise loss function by defining $f$ as $(B \times H \times W \times D) \times C_i$, where $B$ is batch size. Then, let $\mathcal{S}_1$ be a set of feature $f$ from batch $B$. Therefore, we not only considered long dependency in the same volume, but also the intra-dependency of each feature from each volume. Moreover, to acquire robustness against the class imbalance problem, we added a weight for each class. Then, the loss can be defined as:
\begin{equation}
   \mathcal{L}_{feature_1} = - \sum_c^{class} w_c \sum_{\mathcal{P}^c, \mathcal{Z}^c \in \mathcal{S}_1} \mathcal{L}_{voxel-wise}(p, z'),
   \label{eq:6}
\end{equation}
where $z'$ is $stopgrad(z)$ and $w_c$ is ${{|\mathcal{P}| / |\mathcal{P}^c|} \over \sum_c({|\mathcal{P}| / |\mathcal{P}^c|})}$.

\begin{figure}[t]
\begin{center}
    \subfloat[3D U-Net \cite{3DUNet}]{\includegraphics[height=1.3in]{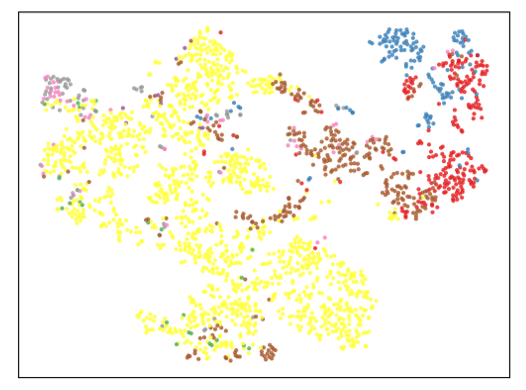}}
    \\
    \subfloat[Ours $\mathcal{L}_{feature_1}$]{\includegraphics[height=1.3in]{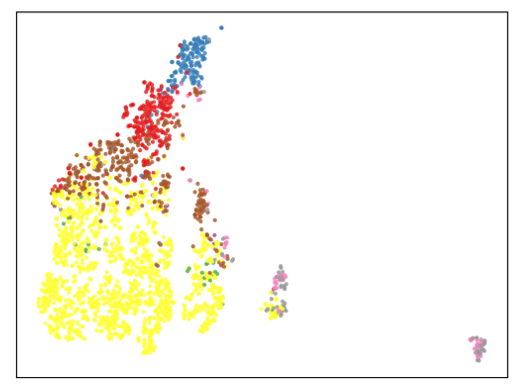}}
    \subfloat[Ours $\mathcal{L}_{feature_2}$]{\includegraphics[height=1.3in]{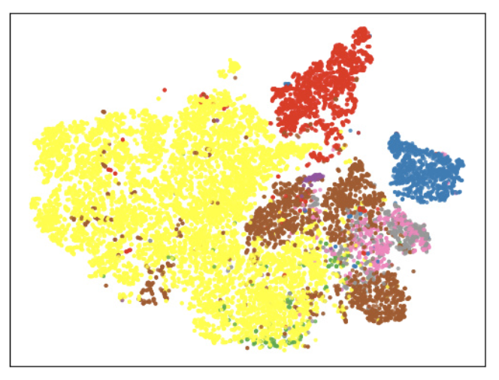}}
    \caption{Visualization of features learned with (a) typical 3D U-Net, (b) our feature loss, $\mathcal{L}_{feature_1}$, and (c) our feature loss, $\mathcal{L}_{feature_2}$. Features are colored according to class labels and we used the test dataset for visualization (test labels are used for only visualization).}
\label{fig:tsne}
\end{center}
\end{figure}

\subsection{Multi-Resolution Context Aggregation}
We introduce the feature loss $\mathcal{L}_{feature_1}$ to suppress intra-voxel-wise relations from the same class. We can enhance the embedding of global context by applying this loss in the last layer of the encoder. However, this method only improves discriminativeness and quality of the last hidden-layer feature map. Moreover, in the segmentation task, learning the local context for accurate localization is also important; however, this loss cannot directly influence the local context. The higher-level hidden layers have relatively larger receptive fields than lower-level hidden layers, which implies that they contain less local context than lower-level layers. When we only consider the features of the last layer, these features have a high receptive field that can ignore local information; further, these features are too downsampled that even the loss of small objects can occur. This is critical for organs such as the pancreas that account for the smallest portion of the 3D volume because the feature relation of the pancreas cannot be defined. As shown in Fig. \ref{fig:tsne} (b), our feature loss is more effective for feature embedding than Fig. \ref{fig:tsne} (a) but it is dominated by a large portion of classes and some of the classes are missing.

Inspired by previous studies \cite{lee2014deeplysupervised, xu2021seed}, we propose multi-resolution context aggregation for obtaining both a highly discriminative quality in the intermediate hidden layers and local context. The method is simple; we only add voxel-level feature layers in the intermediate layers for features $F_2$ and $F_3$. Then, we can define intra-voxel-wise relations from different hidden layers, which is referred to as context aggregation. Based on context aggregation, the semantic information includes not only the global context but also the local context because lower-level hidden layers contain information from the local regions. Furthermore, by enhancing the discriminative quality of the intermediate hidden layers, we also have the advantage of localization. In the 3D U-Net architecture, there is a skip layer that integrates intermediate hidden layer of the encoder and decoder for localization. Learning useful feature information is directly related to the quality of the score map. Thus, this method improves the quality of the segmentation map.

The multi-resolution context aggregation method is important for embedding good semantic information both locally and globally. The final voxel-level feature loss function can be defined as:
\begin{equation}
    \mathcal{L}_{feature_{\mid F \mid}} = - \sum_c^{class} w_c \sum_{\mathcal{P}^c, \mathcal{Z}^c \in \mathcal{S}_{\mid F \mid}} \\ w_f \cdot \mathcal{L}_{voxel-wise}(p, z'),
\label{eq:6}
\end{equation}
where $\mathcal{S}_{\mid F \mid}$ is a set of multi-resolution features, $\mid F \mid$ indicates the total number of hidden layers that is used for defining the feature relation, and $w_f$ is the weight of each hidden layers. Here, we set $\mid F \mid$ as 3, that is, we used the last three hidden layers; moreover, the weights $w_f$ was set as 1.

Finally, we used both voxel-level feature loss and Dice loss for achieving a good representation power and precise segmentation map. Note that the voxel-level feature layer is only used in the training phase, which indicates that no extra capacities are required in the inference phase (i.e., time and space). The total loss can be computed as follows:
\begin{equation}
    \mathcal{L} = \mathcal{L}_{dice} + \lambda \mathcal{L}_{feature_{\mid F \mid}},
\label{eq:7}
\end{equation}
where $\lambda$ is a weighting parameter.

\par

\begin{table*}[h]
\captionsetup{justification=centering, labelsep=newline}
\caption{Quantitative comparison of the proposed methods with other networks in terms of Dice coefficient.}
\label{T:dice-score}
\footnotesize
\begin{center}
\begin{tabular}{| c | c || c  c c  c  c  c  c  c |}
\hline
\textbf{Method} & \multicolumn{9}{ c |}{\textbf{DSC $\uparrow$}}  \\ 
\cline{2-10}
& \textbf{avg}
& \textbf{spleen} & \textbf{left kidney} & \textbf{gallbladder} & \textbf{esophagus}
& \textbf{liver} & \textbf{stomach} & \textbf{pancreas} & \textbf{duodenum}\\
\hline

3D U-Net \cite{3DUNet}& 0.786 & 0.931 & 0.923 & 0.75 & 0.608 & 0.952 & 0.839 & 0.701 & 0.583 \\ \hline
Residual 3D U-Net \cite{ResidualUNet}& 0.784 & 0.928& \textbf{0.937}  & 0.735  & 0.593& 0.954 & 0.823 & 0.714 & 0.591 \\ \hline
VNet \cite{VNet}& 0.773 & 0.923  & 0.913 & 0.766 & 0.601 & 0.943 & 0.814 & 0.668 & 0.555 \\ \hline
Attention U-Net \cite{attention_1}&  0.787 & \textbf{0.944} & 0.932 & 0.726 & 0.588& \textbf{0.956} & 0.846 & 0.719 & 0.584\\ \hline
Supervised Contrastive Loss & 0.787 & 0.914 &  0.935 & 0.704 &0.618 & 0.955 & 0.838 & 0.715 & 0.590 \\ \hline
\textbf{Ours} & \textbf{0.806} & 0.943 & \textbf{0.937} & \textbf{0.793} & \textbf{0.620} & 0.955 & \textbf{0.869} & \textbf{0.725} & \textbf{0.609} \\ \hline

\end{tabular}
\end{center}
\end{table*}

\begin{table}[t]
\captionsetup{justification=centering, labelsep=newline}
\caption{Quantitative comparison of the proposed methods with other networks in terms of parameter size, 95\% Hausdorff distance (HD95) and average symmetric surface distance (ASSD).}
\label{T:params}
\begin{center}
\begin{tabular}{| c || c | c | c |}
\hline
\textbf{Method} & \textbf{\#params} & \textbf{95\% HD $\downarrow$} & \textbf{ASSD $\downarrow$} \\ \hline

3D U-Net \cite{3DUNet}& 16.313M & 3.170 & 0.845 \\ \hline
Residual 3D U-Net \cite{ResidualUNet}& 141.229M & 3.311 & 0.832  \\ \hline
VNet \cite{VNet}& 9.449M & 5.861  & 1.33  \\ \hline
Attention U-Net \cite{attention_1}&  16.836M & 3.013 & 0.812 \\ \hline
Supervised Contrastive Loss & 16.313M & 3.621 &  0.815 \\ \hline
\textbf{Ours} & \textbf{16.313M} & \textbf{2.461} & \textbf{0.681} \\ \hline

\end{tabular}
\end{center}
\end{table}

\section{EXPERIMENTAL RESULTS}

\subsection{Dataset details}
We used 90 abdominal CT images, i.e., 43 from Pancreas-CT and 47 from Beyond the Cranial Valut (BTCV) dataset \cite{DenseVNet}, and referenced standard segmentations of the spleen, left kidney, gallbladder, esophagus, liver, stomach, pancreas, and duodenum. In the dataset, the slice thickness was in the range of 0.5-5.0 mm and pixel sizes were in the range of 0.6-1.0 mm. The dataset was separated into two sets: training and testing, with 70 images for training and 20 for testing. We resampled all abdominal CT images into 128 x 128 x 64. We preprocessed the image using a soft-tissue CT windowing range of [-200, 250] Hounsfield units. After rescaling, we normalized the input images to zero mean and unit variance (i.e., the range of the value is $[0, 1]$).

\subsection{Implementation details}
For training, we used a 3D U-Net \cite{3DUNet} architecture as the base network. The architecture of our voxel-level feature layer was based on a previous study \cite{simsiam}, with the dimensions of all hidden layers set to 64. We used a batch size of four, an Adam optimizer, a weight decay of 0.00001, and a learning rate of 0.001 for our experiment. We adopted a polynomial annealing policy \cite{chen2017rethinking} to schedule the learning rate, which was multiplied by $(1 - {epoch \over total\_epoch})^{p}$ with $p=0.9$. The weighting parameter $\lambda$ was 10 for $\mathcal{L}_{feature_3}$ and 100 for $\mathcal{L}_{featre_1}$. We sampled a maximum of 1700 features from all voxel features for each hidden layer. Further, we sampled more features from false-negative data than true-positive data, where the maximum number of sampled false-negative data was 1000. The network was trained for 500 epochs. We implemented our framework in PyTorch \cite{paszke2019pytorch}, using an NVIDIA TITAN RTX GPU. At the inference time, only the 3D U-Net network was used for segmentation.

\subsection{Results}
For our evaluation metrics, we used Dice score coefficient (DSC) \cite{dice_loss}, Hausdorff distance (HD95; mm) \cite{hd, hd2}, and average symmetric surface distance (ASSD; mm) \cite{assd}. HD95 is considered as a better-generalized evaluation metric for distance because of the existence of ground-truth variations (e.g., portal vein regions adjacent to the liver). In our experiments, we evaluated the performance in terms of accuracy of our proposed network by comparing the results with those of the state-of-the-art models, i.e., 3D U-Net \cite{3DUNet}, Residual 3D U-Net \cite{ResidualUNet}, VNet \cite{VNet}, attention U-Net \cite{attention_1}, and supervised contrastive loss (SCL). For a fair comparison, we did not perform any postprocessing for the output. SCL is also based on the 3D U-Net architecture and used a typical contrastive learning method for voxel-level feature learning, as discussed in \cite{simclr, wang2021exploring}. The SCL method used the same multi-resolution layer feature and number of sampling positive features as in $L_{feature_3}$, and the number of negative samples was set as 1700 for each layer.

\begin{figure}[t]
\begin{center}
    \includegraphics[height=0.2in]{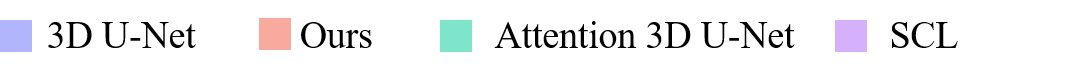}\\
    \subfloat[spleen]{\includegraphics[height=0.35in]{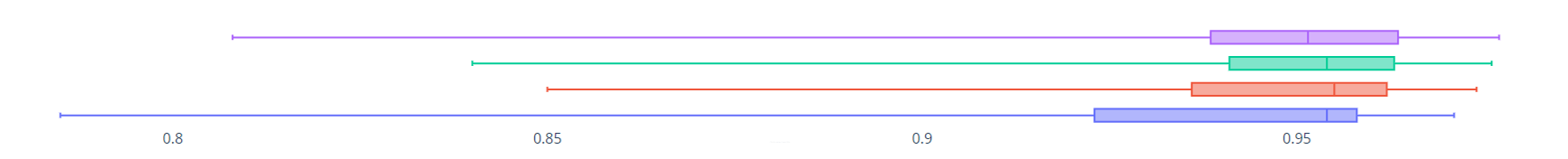}}\\
    \subfloat[left kidney]{\includegraphics[height=0.35in]{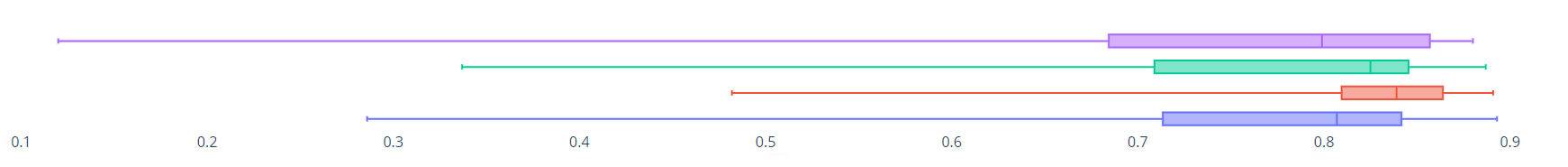}}\\
    \subfloat[gallbladder]{ \includegraphics[height=0.35in]{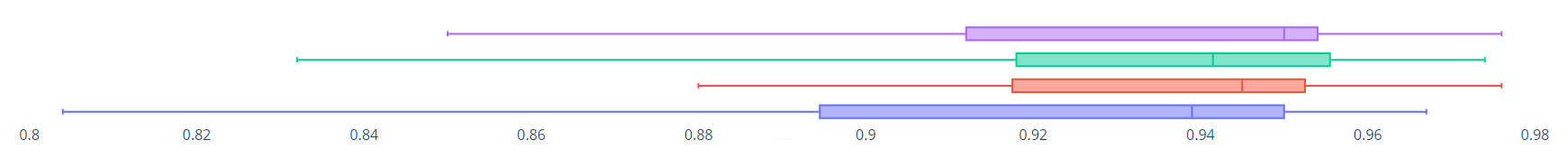}}\\
    \subfloat[esophagus]{ \includegraphics[height=0.35in]{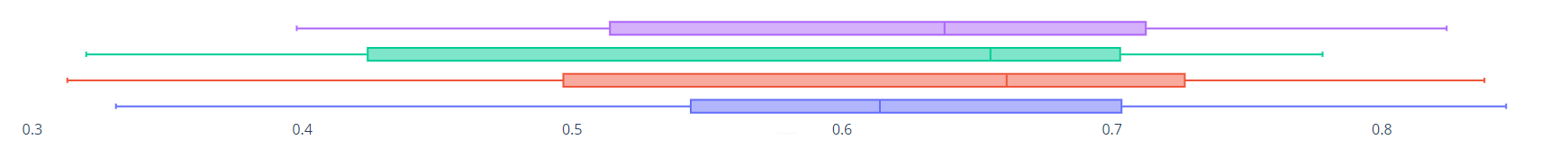}}\\
    \subfloat[liver]{ \includegraphics[height=0.35in]{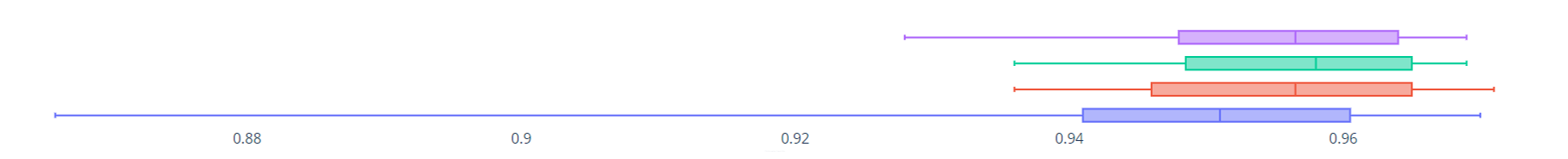}}\\
    \subfloat[stomach]{ \includegraphics[height=0.35in]{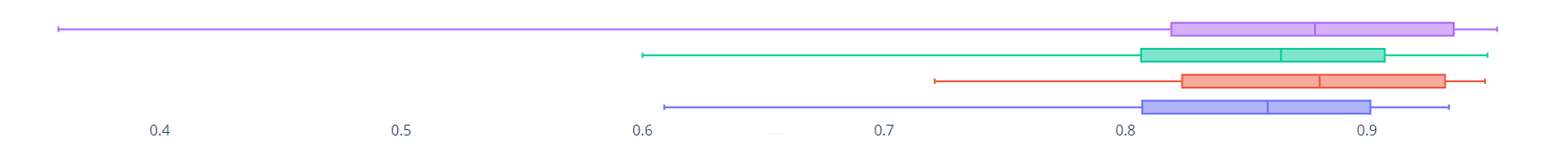}}\\
    \subfloat[pancreas]{ \includegraphics[height=0.35in]{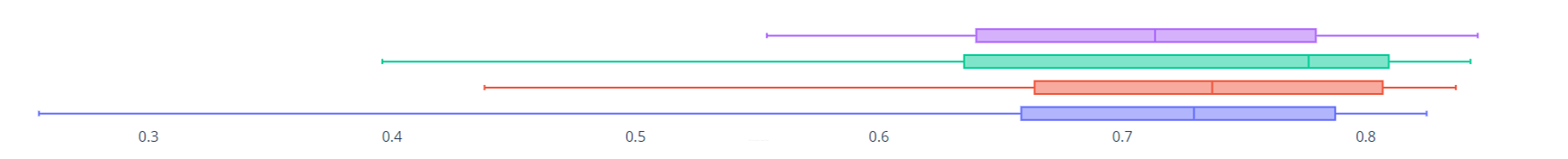}}\\
    \subfloat[duodenum]{ \includegraphics[height=0.35in]{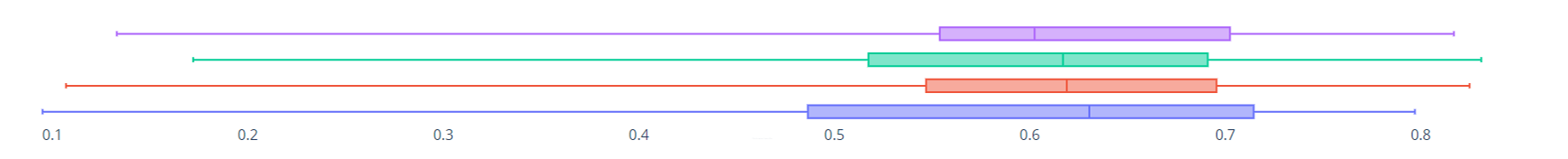}}
    \caption{Box plots of the Dice score coefficient (DSC) of eight different organs for different approaches.}
    \label{fig:plot}
\end{center}
\end{figure}

\begin{figure*}[t]
\begin{center}
    \includegraphics[height=0.3in]{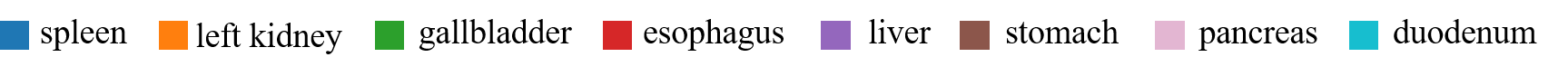}
    \subfloat[GroundTruth]{\includegraphics[height=2.2in]{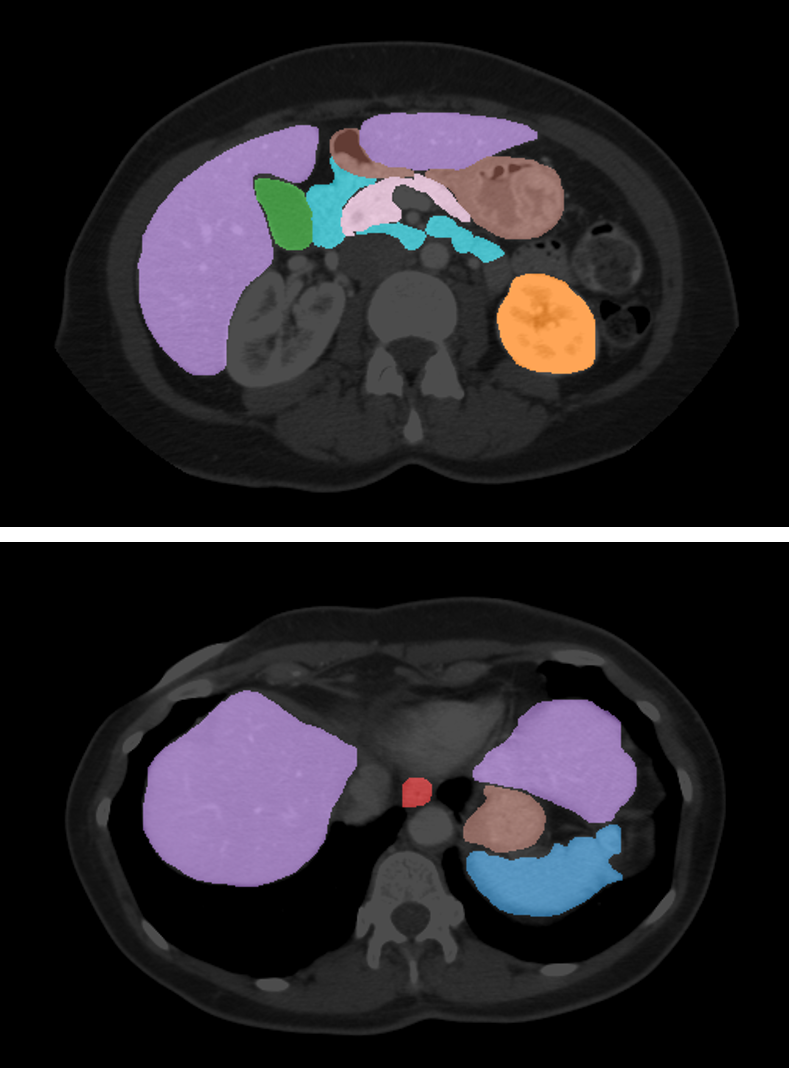}}
    \hspace{0.01in}
    \subfloat[Ours]{\includegraphics[height=2.2in]{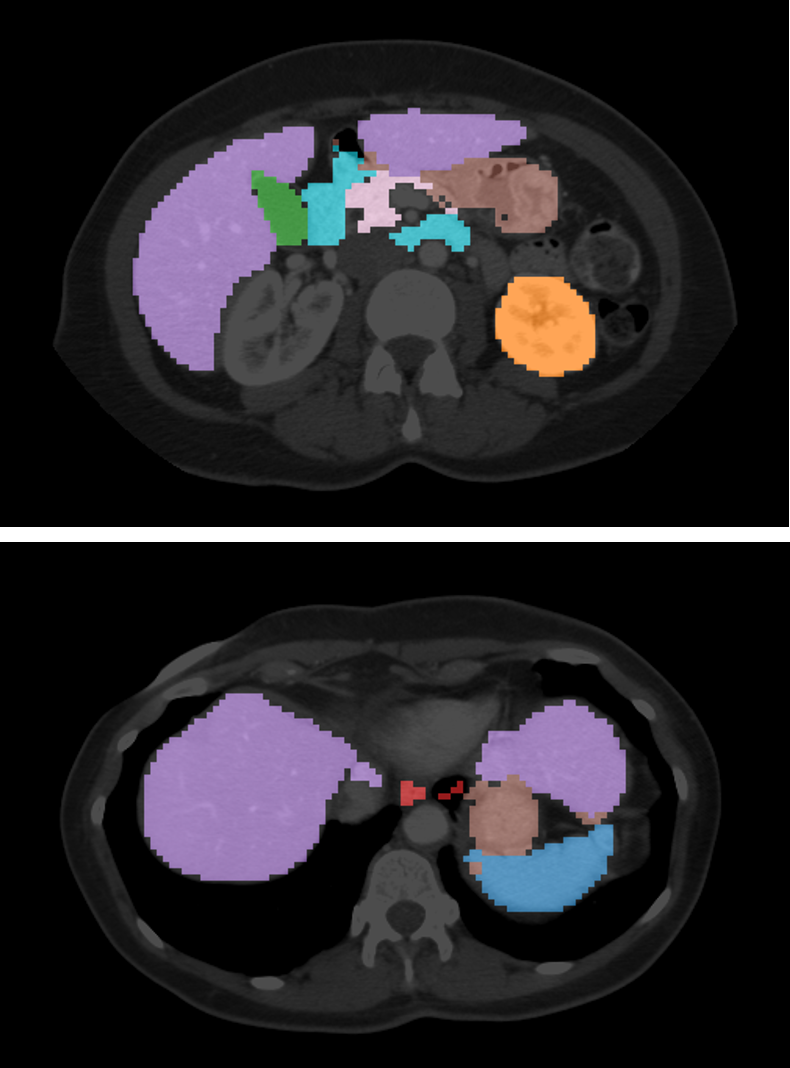}}
    \hspace{0.01in}
    \subfloat[SCL]{ \includegraphics[height=2.2in]{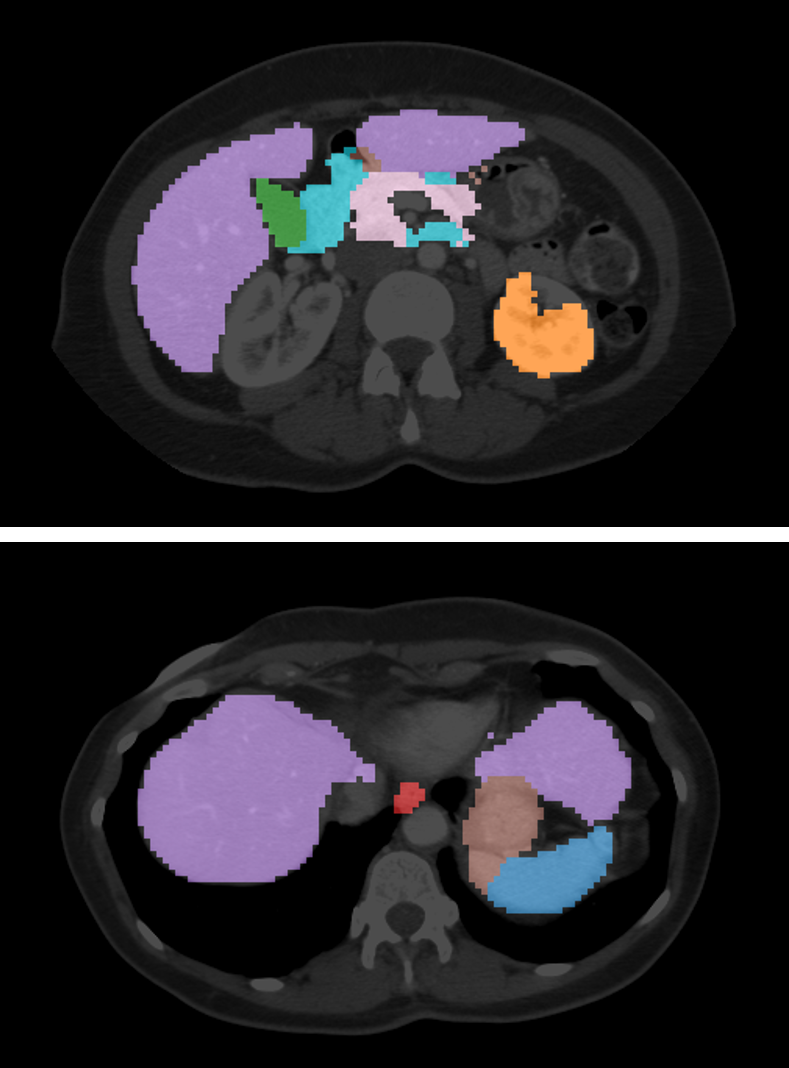}}
    \hspace{0.01in}
    \subfloat[Attention U-Net \cite{attention_1}]{ \includegraphics[height=2.2in]{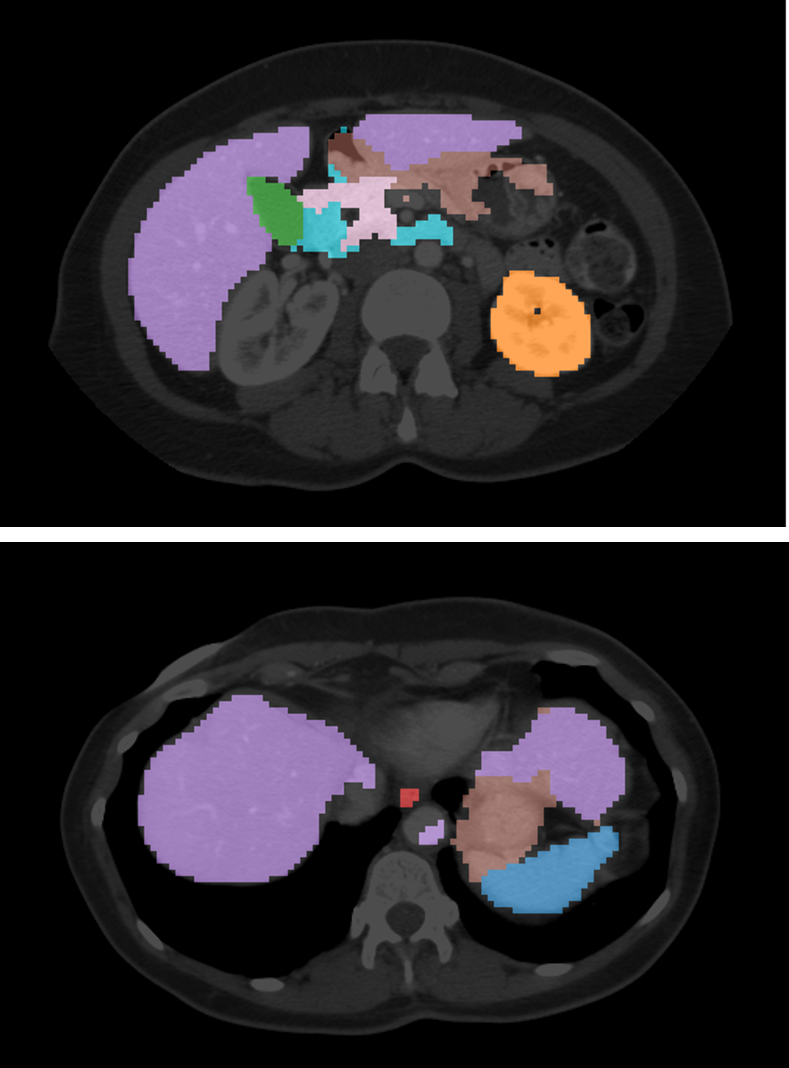}}
    \caption{Qualitative comparison of different approaches by 2D visualization. From left to right: (a) GroundTruth, (b) Our proposed method, (c) Supervised Contrastive Loss (SCL), (d) Attention 3D U-Net \cite{attention_1}.}
    \label{fig:2d_vis}
\end{center}
\end{figure*}

\begin{figure*}[t]
\begin{center}
    \subfloat[GroundTruth]{\includegraphics[height=1.2in]{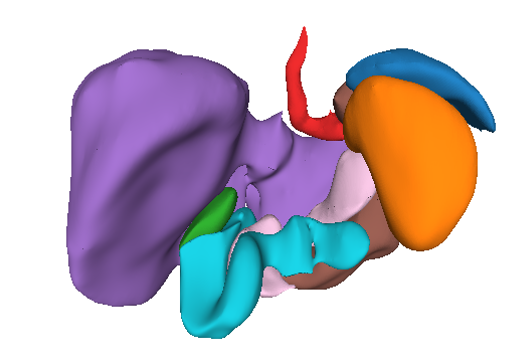}}
    \subfloat[Ours]{\includegraphics[height=1.2in]{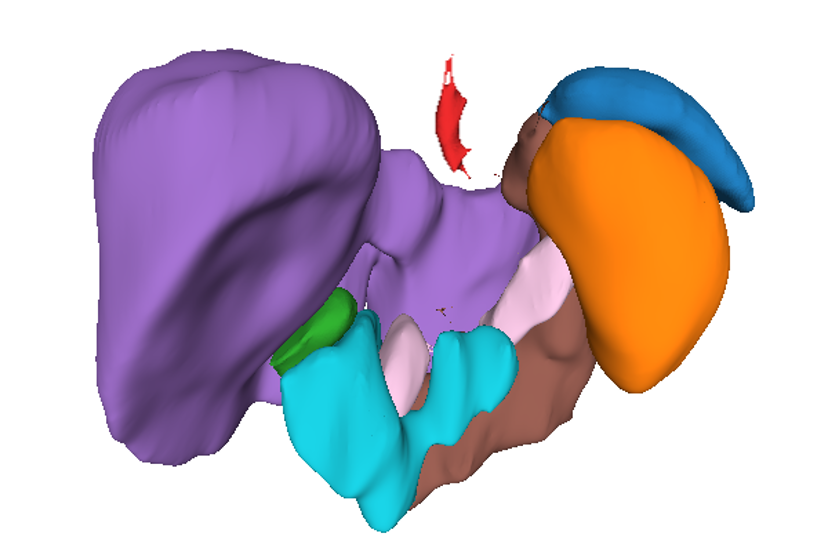}}
    \subfloat[SCL]{ \includegraphics[height=1.2in]{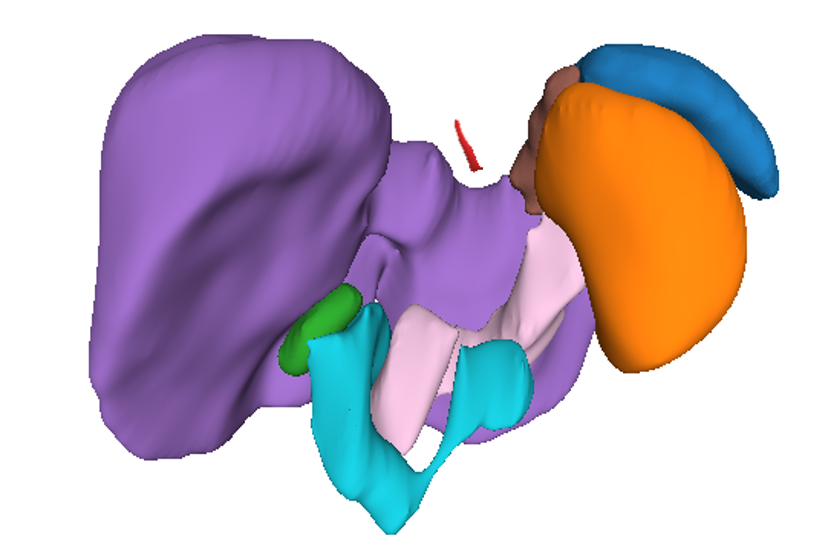}}
    \subfloat[Attention U-Net \cite{attention_1}]{ \includegraphics[height=1.2in]{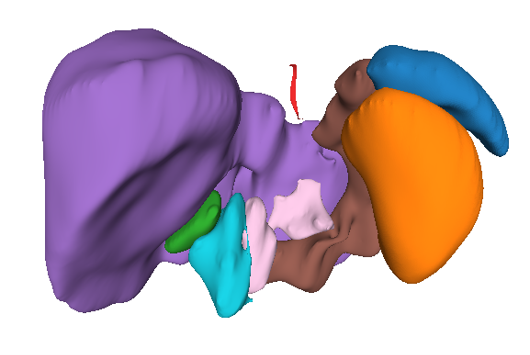}}
    \caption{Qualitative comparison of different approaches by 3D visualization. From left to right: (a) GroundTruth, (b) Our proposed method, (c) Supervised Contrastive Loss (SCL), (d) Attention 3D U-Net \cite{attention_1}.}
    \label{fig:3d_vis}
\end{center}
\end{figure*}

\textbf{Quantitative results} Table \ref{T:dice-score} and Table \ref{T:params} list the quantitative results of multi-organ segmentation. The results show that our proposed method outperformed previous methods with significant improvements in all evaluation metrics. We achieved superior values of DSC (0.806), HD95 (2.461) and ASSD (0.681) with the same number parameters as the 3D U-Net \cite{3DUNet}. VNet \cite{VNet} performed the worst when compared to the other methods. Our proposed method outperformed attention U-Net \cite{attention_1} in terms of all evaluation metrics; moreover, our method used fewer parameters when compared to \cite{attention_1}. The SCL, which also uses voxel-level feature learning and employs contrastive loss, showed a small improvement when compared to 3D U-Net \cite{3DUNet}. The relatively inferior performance of SCL was because the contrastive loss required a large number of negative samples and a large batch size, which is difficult to obtain in 3D volumetric applications. From Table \ref{T:dice-score}, it can be observed that our proposed method achieved significant improvements in the segmentation of the gallbladder, esophagus, pancreas and duodenum when compared to other organs. Specifically, our method showed superior performance for small organs. Box plots of the results listed in Table \ref{T:dice-score} are illustrated in Fig. \ref{fig:plot}.
 
\textbf{Qualitative results} Our qualitative results are illustrated in Fig. \ref{fig:2d_vis} (2D visualization) and \ref{fig:3d_vis} (3D visualization). We chose top-two related methods for comparison, i.e., SCL and attention U-Net \cite{attention_1}. When compared to other methods, our results have a higher overlap ratio with respect to GroundTruth (i.e., stomach and duodenum). Figure \ref{fig:3d_vis} illustrates that the stomach is difficult to detect in SCL, and the duodenum is partially missing in attention U-Net \cite{attention_1} when compared to our method. Moreover, our method showed fewer false-positive responses for the duodenum and pancreas (Fig. \ref{fig:3d_vis}). 

\subsection{Ablation Study}

\begin{figure}[!t]
\begin{center}
\begin{tikzpicture}[thick,scale=0.8, every node/.style={scale=0.8}]
    \begin{axis}[
        xlabel=\% of labels,
        ylabel=DSC, legend pos=outer north east]
    \addplot[smooth,mark=*,blue] plot coordinates {
        (10,0.431)
        (20,0.609)
        (50,0.748)
        (100,0.786)
    };
    \addlegendentry{3D U-Net}

    \addplot[smooth,color=red,mark=*]
        plot coordinates {
            (10,0.548)
            (20,0.674)
            (50,0.764)
            (100,0.806)
        };
    \addlegendentry{Ours}
    \end{axis}
\end{tikzpicture}
\end{center}
\caption{DSC evaluation of the proposed method with different percentage of labels.}
\label{plot}
\end{figure}
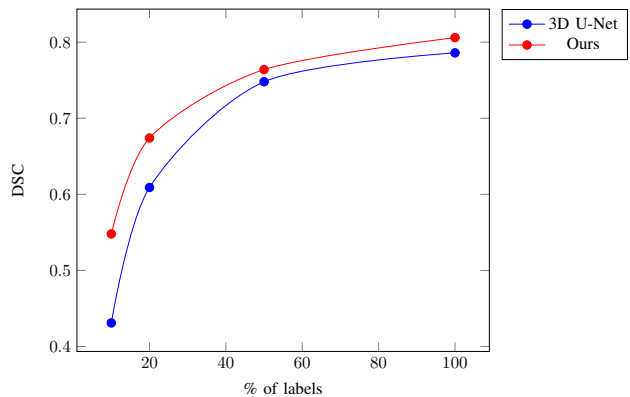

We investigated the effectiveness of the different components in our proposed method. Table \ref{T:ablation1}-\ref{T:ablation3} and Fig. \ref{plot} show the quantitative results (DSC, HD95 and ASSD) of different model settings. We conducted experiments to verify the effectiveness of our newly proposed methods: multi-resolution context aggregation, loss functions, and different architectures. Furthermore, we performed extensive experiments in terms of the size of the training dataset. The examinations are described in detail in the following subsections.

\textbf{Multi-Resolution Context Aggregation} We first analyzed the impact of our multi-resolution context aggregation. First, when we compared representation embeddings of feature using $\mathcal{L}_{feature_1}$ and $\mathcal{L}_{feature_2}$ as shown in Fig. \ref{fig:tsne}; based on this comparison, it was observed that feature embedding where context aggregation from multi-layers is considered as more structured than that of a single layer. In Table \ref{T:ablation1}, we list the quantitative segmentation results using three different proposed feature losses, where the first row lists the results of 3D U-Net trained with $\mathcal{L}_{feature_1}$ and the second and third rows list the results with $\mathcal{L}_{feature_2}$, $\mathcal{L}_{feature_3}$, respectively. The first row is similar to recent works on contrastive learning as it applies loss in the last layer of the encoder. Then, we aggregated the contexts of the last two hidden layers, which is listed in the second row, showing performance improvement when compared to the first row. Moreover, aggregating the contexts of last three hidden layers significantly improved performance, which can be explained by the fact that aggregating the local and global contexts results in a more precise segmentation output.

\textbf{Weighted Loss Function} We also compared our loss function with and without a weighted function (i.e., \ref{eq:6}). Table \ref{T:ablation2} lists our training result obtained by learning using the $\mathcal{L}_{feature_3}$ function. The first row lists the results when the weighted function is not applied, and the second row lists the results of the final loss function. The feature loss without weighted loss function even showed superior performance when compared other methods in terms of DSC and HD95. Meanwhile, applying weighted function improved the performance itself and stability of training.

\begin{table}[t]
\captionsetup{justification=centering, labelsep=newline}
\caption{Ablation study of the multi-resolution context aggregation in terms of Dice score coefficient (DSC), HD95 and ASSD.}
\label{T:ablation1}
\begin{center}
\begin{tabular}{| c || c | c | c |}
\hline
\textbf{Method} & \textbf{DSC} & \textbf{HD95} & \textbf{ASSD} \\ \hline

feature (1) & 0.793 & 3.210 & 0.86 \\ \hline
feature (2) & 0.797 & 2.895 & 0.761   \\ \hline
feature (3) & \textbf{0.806} & \textbf{2.461} &  \textbf{0.681} \\ \hline

\end{tabular}
\end{center}
\end{table}

\begin{table}[t]
\captionsetup{justification=centering, labelsep=newline}
\caption{Ablation study of the weighted loss in terms of DSC, HD95 and ASSD.}
\label{T:ablation2}
\begin{center}
\begin{tabular}{| c || c | c | c |}
\hline
\textbf{Method} & \textbf{DSC} & \textbf{HD95} & \textbf{ASSD} \\ \hline

feature (w/o weight) & 0.8 & 2.918 &  0.825 \\ \hline
feature & \textbf{0.806} & \textbf{2.461} &  \textbf{0.681} \\ \hline

\end{tabular}
\end{center}
\end{table}

\begin{table}[t]
\captionsetup{justification=centering, labelsep=newline}
\caption{Ablation study of the hidden dimension in terms of DSC, HD95 and ASSD.}
\label{T:ablation3}
\begin{center}
\begin{tabular}{| c || c | c | c |}
\hline
\textbf{Method} & \textbf{DSC} & \textbf{HD95} & \textbf{ASSD} \\ \hline

feature (64) & 0.793 & 3.210 & 0.86 \\ \hline
feature (128) & 0.793 &  3.263 & 0.822  \\ \hline
feature (256) & \textbf{0.796} & \textbf{3.089} & \textbf{0.804}  \\ \hline

\end{tabular}
\end{center}
\end{table}

\textbf{Hidden Dimension} Then, we analyzed the efficacy of the SimSiam method for different hidden dimensions. In our projection and prediction layers in the voxel-level feature layer, MLP heads are used for feature projection and prediction. It is important to choose an appropriate number of hidden dimensions. Table \ref{T:ablation3} lists the results of the experiments on different hidden dimensions (64, 128 and 256). When the number of hidden dimensions was 64 or 128, the experiments achieved only minimal improvements in HD95 but showed worse results in ASSD. When the number of hidden dimensions was 256, superior results were achieved in all metrics, which proves that superior performance can be obtained by not reducing the original feature dimension.

\textbf{Dataset Size} We investigated our proposed method with different percentage of labels. Our method is an efficient representation learning method for embedding rich information with limited annotated data. We compared our method with the original 3D U-Net network with a limited dataset size. As shown in Fig. \ref{plot}, by using a small dataset size (10 and 20\%), we achieved a significant improvement in DSC (i.e., 0.431 (3D U-Net) to 0.548 (ours)). This shows that simply adding our methods to a base network is efficient for learning generalized and high-level representations. 

\par

\section{DISCUSSION}

Recent studies on multi-organ segmentation have explored various deep learning architectures, such as encoder-decoder based networks \cite{VoxelResNet, 3DUNet, ResidualUNet, VNet} or attention networks \cite{attention_1, attention_2}, for encoding high-level features using limited training data. However, training the network in the final output space makes it difficult to embed rich global and local contextual features in the representation space (as shown in Fig. \ref{section:3-b} and Figure \ref{fig:tsne}). Our proposed method is effective for representation learning in that it suppresses voxel-wise relations in the representation space, i.e., it enables voxels from the same class to be projected to the same point in the representation space. Moreover, by learning voxel-wise feature relations based on the SimSiam \cite{simsiam} method, the utilization of negative samples can be avoided; consequently, our method is superior in terms of computational efficiency. A recent work \cite{wang2021exploring}, which employed contrastive loss for supervised segmentation tasks, showed minimal improvement on our dataset (Table \ref{T:dice-score}). Because it is difficult to consider a large amount of negative samples in 3D volumetric datasets, it is challenging to employ existing contrastive learning methods in medical imaging applications. Our proposed method, which does not require negative samples during training, suggests a simple way of employing contrastive loss in the medical image segmentation task. Furthermore, by introducing a multi-resolution context aggregation method, we encoded both local and global contexts in the representation space. We achieved a more structured representation space (Fig. \ref{fig:tsne}) and precise segmentation results (Fig. \ref{fig:2d_vis}) when compared to the previous method \cite{wang2021exploring} that only considered the global context. Our extensive experiments demonstrated that our method provides more informative feature embedding, resulting in superior accuracy, especially with a limited data size (Fig. \ref{fig:plot}). Moreover, our method can be simply added to existing networks, which implies that the proposed metric can be easily extended to other dense prediction tasks. In the future, our method can be improved by developing more efficient feature sampling techniques or new representation losses.

\section{CONCLUSION}

In this work, we proposed a new voxel-level representation learning method for multi-organ segmentation task. Our method enforces voxel-wise relations while preserving computational efficiency and encodes both the local and global contexts for achieving precise segmentation results. Our method successfully performed the rich representation learning by employing a 3D U-Net architecture as a backbone without introducing additional parameters. The experimental results demonstrated that our method is superior to other contrastive loss-based methods. The proposed method is also independent of the model architecture, which indicates that our algorithm can be applied to any types of network models.





\bibliographystyle{plain}
\bibliography{mybib}

\end{document}